\documentclass[runningheads]{llncs}
\usepackage[T1]{fontenc}
\usepackage{graphicx}
\usepackage{makeidx}  
\usepackage[colorlinks,linkcolor=blue]{hyperref}
\usepackage{url}
\usepackage{longtable}
\usepackage{multirow}
\usepackage{amssymb, amsmath, bm}
\usepackage{amsfonts, amssymb}
\usepackage{pifont}
\usepackage{mathtools}
\usepackage{mathrsfs}
\usepackage{booktabs}
\usepackage{cite}
\usepackage{dsfont}
\usepackage{makecell}
\usepackage[noend]{algpseudocode}
\usepackage{algorithmicx,algorithm}
\usepackage[misc]{ifsym} 
\usepackage{bbding}
\usepackage{ulem}
\usepackage[utf8]{inputenc}
\newcommand{\R}{RL-CAS}

\begin{document}

\title{Subtyping Breast Lesions via Generative Augmentation based Long-tailed \\ Recognition in Ultrasound}
\titlerunning{BreastGenAug}
\author{Shijing Chen\inst{1,2,3}\thanks{Shijing Chen and Xinrui Zhou contribute equally to this work.} \and Xinrui Zhou\inst{1,2,3\star} \and Yuhao Wang\inst{1,2,3} \and Yuhao Huang\inst{1,2,3} \and Ao Chang\inst{1,2,3} \and Dong Ni\inst{1,2,3} \and Ruobing Huang\inst{1,2,3}\textsuperscript{(\Letter)}}


\institute{
\textsuperscript{$1$}National-Regional Key Technology Engineering Laboratory for Medical Ultrasound,
School of Biomedical Engineering, Medical School, Shenzhen University, China\\
\email{ruobing.huang@szu.edu.cn}\\
\textsuperscript{$2$}Medical Ultrasound Image Computing (MUSIC) Lab, Shenzhen University, China\\ 
\textsuperscript{$3$}Marshall Laboratory of Biomedical Engineering, Shenzhen University, China}
\authorrunning{S. Chen and X. Zhou et al.}

\maketitle           
\begin{abstract}
Accurate identification of breast lesion subtypes can facilitate personalized treatment and interventions. Ultrasound (US), as a safe and accessible imaging modality, is extensively employed in breast abnormality screening and diagnosis. However, the incidence of different subtypes exhibits a skewed long-tailed distribution, posing significant challenges for automated recognition. Generative augmentation provides a promising solution to rectify data distribution. Inspired by this, we propose a dual-phase framework for long-tailed classification that mitigates distributional bias through high-fidelity data synthesis while avoiding overuse that corrupts holistic performance. The framework incorporates a reinforcement learning-driven adaptive sampler, dynamically calibrating synthetic-real data ratios by training a strategic multi-agent to compensate for scarcities of real data while ensuring stable discriminative capability. Furthermore, our class-controllable synthetic network integrates a sketch-grounded perception branch that harnesses anatomical priors to maintain distinctive class features while enabling annotation-free inference. Extensive experiments on an in-house long-tailed and a public imbalanced breast US datasets demonstrate that our method achieves promising performance compared to state-of-the-art approaches. More synthetic images can be found at \url{https://github.com/Stinalalala/Breast-LT-GenAug}.
\keywords{Breast ultrasound \and Histological subtype \and Long-tailed recognition \and Diffusion model}
\end{abstract}

\section{Introduction}
\label{sec:intro}
\begin{figure*}[!t]
    \centering
    \includegraphics[width=0.93\linewidth]{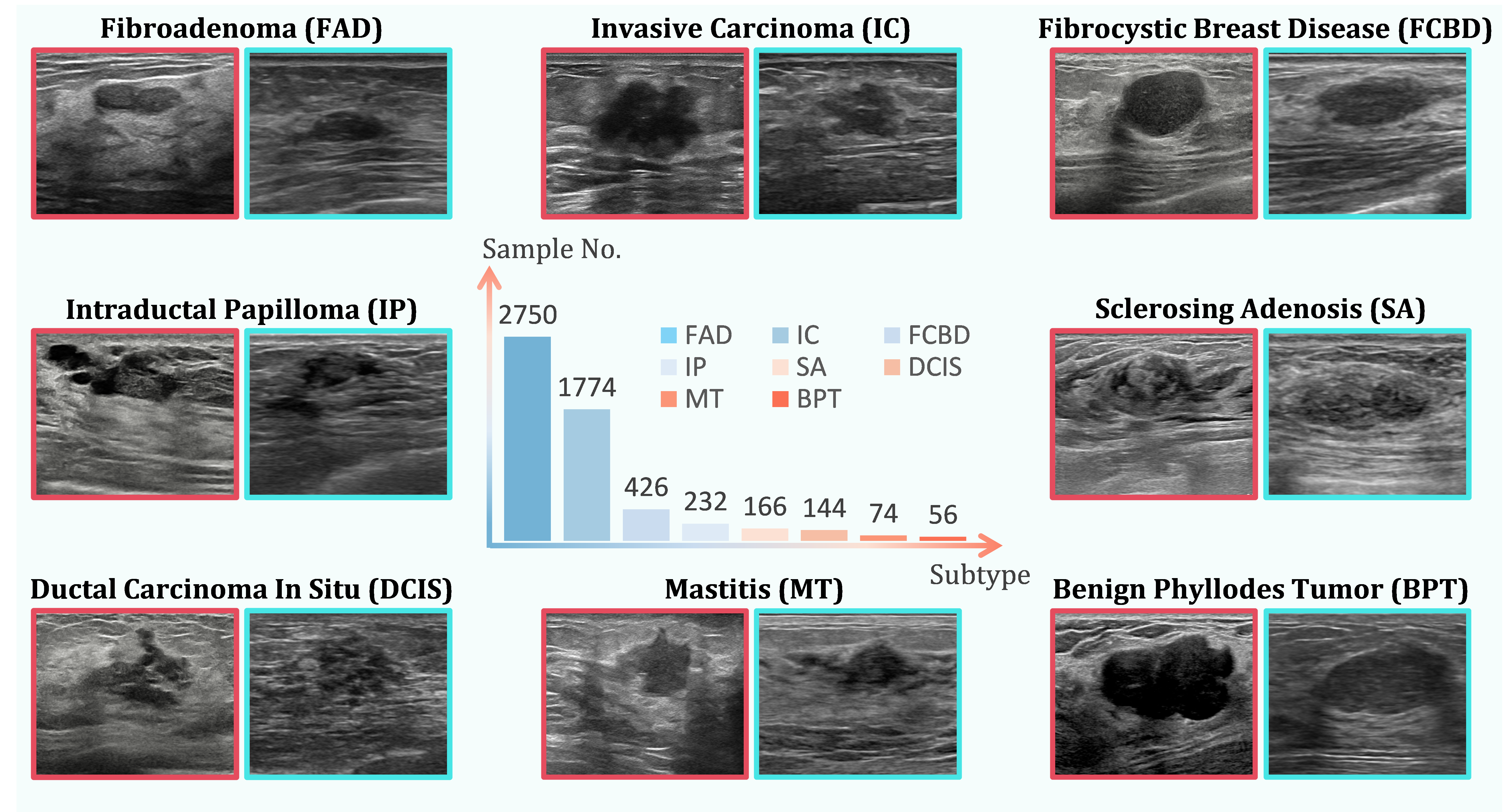}
    \caption{Breast US images of lesions with different histological subtypes. The histogram indicates the incidence of different subtypes, which exhibit a long-tailed distribution. Red-bordered images are real US images, while the blue-bordered ones are synthetic data generated using the proposed framework.}
    \label{fig:Intro Fig1}
\end{figure*}

The global prevalence of breast cancer~\cite{2021Global} drives widespread adoption of early screening tools, with ultrasound (US) emerging as the preferred modality for younger cohorts with dense breasts~\cite{US2017}. 
While standard protocols predominantly emphasize binary lesion classification regarding malignancy/benignity, contemporary clinical findings suggested that the identification of different lesion subtypes may be critical for optimizing treatment planning and management~\cite{barrios2022global}.

However, the incidence rates of different subtypes naturally exhibit an extremely skewed long-tailed distribution that presents substantial challenges for accurate recognition (see Fig.~\ref{fig:Intro Fig1}). This diagnostic complexity is compounded by inter-class morphological overlap (e.g., FAD vs. BPT, IP vs. DCIS in Fig.~\ref{fig:Intro Fig1}), coupled with marked intra-class heterogeneity~\cite{makki2015diversity}. 
Common deep learning classifiers~\cite{becker2018classification,han2017deep,moon2020computer}, despite their advancements, may suffer severe performance degradation in long-tail scenarios.

In recent years, many studies have been proposed for long-tailed recognition~\cite{zhang2023deep}, which can be mainly divided into three categories: 1) Re-balancing-based methods seek to re-balance the negative influence brought by the class distribution asymmetry~\cite{cui2019classbalancedlossbasedeffective,menon2020long}; 2) Architecture-enhanced approaches often ensemble multiple models or incorporate additional modules to improve the robustness against long-tailed data~\cite{shi2024longtaillearningfoundationmodel,du2024probabilistic,li2022nested}; 3) Augmentation-based schemes aim to improve both the quantity and diversity of training datasets. Classical methods apply predefined transformations to data samples or features~\cite{du2023global}, while some recent approaches investigated the potential of applying denoising diffusion probabilistic models (DDPMs) to synthesize high-quality medical images for downstream tasks~\cite{luo2024measurement,Parapat2024generating}. Compared to re-balancing methods, synthetic approaches can rectify data distributions through conditional generation, thereby avoiding both overfitting to head classes and overcompensating for tail classes. Furthermore, as these methods only affect the training phase, they enable fast and lightweight inference during testing without requiring complex network structures in architecture-enhanced models. Nevertheless, current synthetic solutions may still face two limitations in addressing the proposed task: 
1) The predominant paradigm fails to explicitly address distribution skewness and remains susceptible to biased sampling during model training. 
2) Limited capability in capturing fine-grained tissue patterns that may undermine the discriminative power of synthetic instances, particularly in tail classes with limited exemplars.

Based on the above analysis, we propose a dual-phase framework for long-tailed classification that mitigates data asymmetry through high-fidelity data synthesis while dynamically modulates synthetic usage to maintain balanced classification performance. This cascaded data curation pipeline enhances diversity expansion while avoiding noise amplification. It is equipped with a reinforcement learning (RL)-driven class adaptive sampler that automates batch composition by learning to balance head-class stability and tail-class exploration. Additionally, our class-controllable synthetic network is also guided with a sketch-grounded perception branch that injects anatomical priors to retain class-discriminative traits while facilitating annotation-free inference.

\section{Methodology}
\label{sec:method}
\begin{figure*}[!t]
    \centering
    \includegraphics[width=1.0\linewidth]{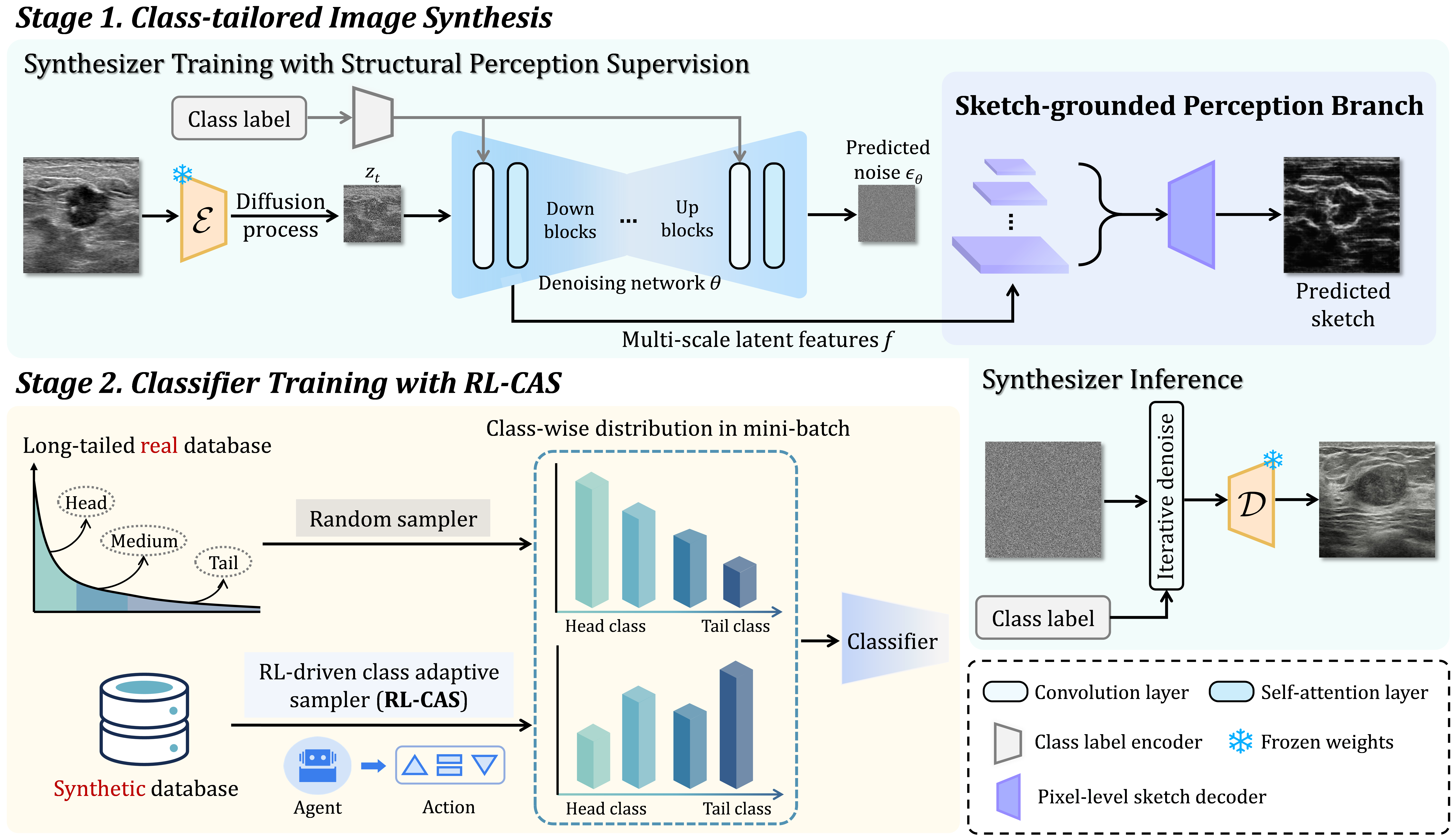}
    \caption{Pipeline of our proposed framework. $\mathcal{E}, \mathcal{D}$ indicate pre-trained encoder and decoder in VAE, respectively. $\mathit{z}_t$ refers to latent features after the $\mathit{t}$-step diffusion process. RL, reinforcement learning.} 
    \label{fig:framework}
\end{figure*}

Fig.~\ref{fig:framework} presents our two-stage framework for long-tailed breast lesion classification in US images. 
The first stage employs a label-conditioned synthesizer with structural perception constraints, enabling class-tailored image synthesis from Gaussian noise while maintaining diagnostic structural fidelity.
The second stage incorporates an RL-driven class adaptive sampler (RL-CAS) that automatically optimizes batch composition by dynamically balancing synthetic-real data during classifier training, effectively addressing long-tailed distribution challenges through focal learning. 

\subsection{Synthesizer for High-fidelity Image Generation}
\textbf{Preliminaries of DDPMs.}
DDPMs~\cite{sohl2015deep,ho2020denoising} are generative models that learn to model data distributions by iteratively denoising corrupted inputs. 
To alleviate the computational burden of pixel-space training in standard DDPMs, latent diffusion models (LDMs)~\cite{rombach2022high} were introduced for perceptual compression.
The objective function of LDM can be formulated as: $\mathcal{L}_{LDM} = \mathbb{E}_{\mathit{z}_0, \epsilon, \mathit{t}, \mathbf{c}}||\epsilon - \epsilon_\theta(\mathit{z}_t, \mathit{t}, \mathbf{c})||^2$, where $\mathit{z}_0$ is the latent code of the training data from a pre-trained variational autoencoder (VAE)~\cite{esser2021taming}, \textit{t} represents the time step, $\epsilon_\theta$ and $\epsilon$ are the predicted and target noise, respectively.
Here, $\mathbf{c}$ indicates the (optional) control signal that the model can be conditioned on.
In the proposed synthesizer, $\mathbf{c}$ is specifically defined as the class label representing the subtype of breast lesions. \\
\textbf{Basic Architecture of Class-steerable Synthesizer.}
Building upon the LDM, our synthesizer architecture integrates two key synergistic components: a. variational latent encoding and b. guided denoising. The VAE encoder compresses medical images into compact latent representations, while the UNet-shaped denoiser progressively removes the noise added to the latent features through several convolution-attention hybrid blocks. To ensure diagnostic relevance, we implement disease-specific generation control by injecting class labels into the denoising trajectory, following a similar approach in~\cite{zhou2024ctrl}. \\
\textbf{Structural Perception Supervision for Refined Reconstruction.}
Label-guided data synthesis achieves basic class matching but struggles with anatomical structure fidelity in complex cases, limiting its practical utility. Current methods often inject geometric conditions (e.g., segmentation masks) in the denoiser to improve structural control~\cite{zhang2023adding,mou2024t2i}.
However, such implicit conditioning may degrade generative capacity in the absence of test-time annotation inputs, which is commonly unavailable in clinical scenarios~\cite{zhou2024heartbeat}.
Inspired by ~\cite{wu2023datasetdm,wang2024detdiffusion}, we propose to overcome this by introducing explicit structural perception supervision through sketches to capture anatomical priors.
This design enables annotation-free inference while preserving fine anatomical details.

To this end, as shown in Fig.~\ref{fig:framework}, we introduce a sketch-grounded perception branch during synthesizer training.
Specifically, multi-scale latent features $f = [f_1, f_2, f_3, f_4]$ from the encoding path of the denoising network, corresponding to resolutions of $[H\times W, \frac{H}{2}\times\frac{W}{2}, \frac{H}{4}\times\frac{W}{4}, \frac{H}{8}\times\frac{W}{8}]$, are extracted as the branch input.
$H,W$ represent the height and width of the latent features before input to the denoiser, respectively.
Then, we propose a pixel-level sketch decoder that concatenates $f$ and performs upsampling via transposed convolution to produce sketch predictions.
To achieve refined structural reconstruction, this study introduces a customized sketch loss, utilizing the rich anatomical priors from sketches.
The core idea is to optimize the high-dimensional feature space of the denoiser in a self-supervised manner through an $L_1$ loss to minimize the reconstruction error between the predicted sketch $S_{pred}$ and its ground truth counterpart $S_{gt}$.
Note that $S_{gt}$ is precomputed using the sketch extractor~\cite{su2021pixel}.
To balance the noise scale in the latent features, referring to~\cite{wang2024detdiffusion}, we apply $\sqrt{\bar{\alpha}_t}$ from DDPM~\cite{ho2020denoising} as scaling factors to the above $L_1$ loss for ensuring feature efficacy and easing the synthesizer training.
To summarize, our sketch loss for structural perception supervision can be described as: $\mathcal{L}_{s}=\sqrt{\bar{\alpha}_t}\cdot L_1(S_{pred},S_{gt})$.
This encourages the synthesizer to focus on latent features with lower noise (i.e., smaller time steps) and vice versa.
Eventually, the objective function of our synthesizer combines the basic $\mathcal{L}_{LDM}$ with $\mathcal{L}_{s}$, which is represented as follows: $\mathcal{L}=\mathcal{L}_{LDM}+\lambda\mathcal{L}_{s}$, where $\lambda$ is a hyperparameter that controls the strength of the perception supervision.

\begin{figure*}[!t]
    \centering
    \includegraphics[width=1.0\linewidth]{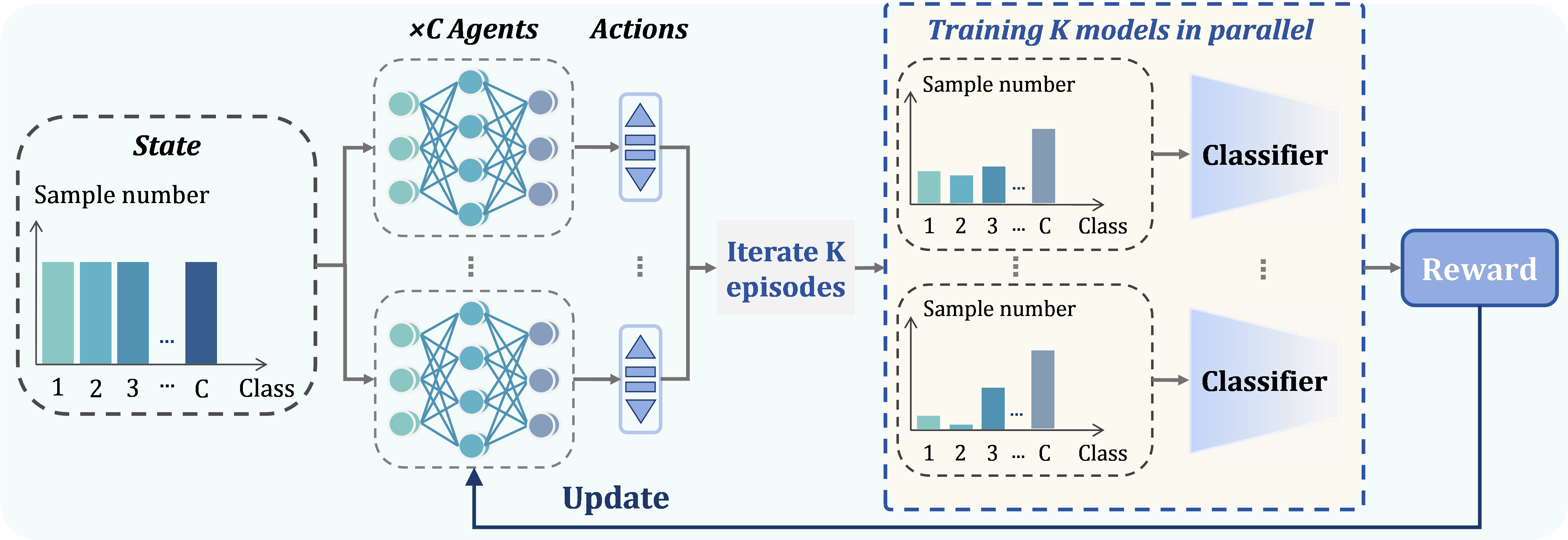}
    \caption{Illustration of the RL-driven class adaptive sampler. Starting from a uniform initial state (green histogram on the left), multiple agents take different actions to modify the ratio of synthetic data during training to optimize the reward.  } 
    \label{fig:online_samler}
\end{figure*}

\subsection{RL-driven Class Adaptive Sampler for Enhanced Classification}
Strategic deployment of these synthesized data during downstream training is equally important for effective long-tailed classification, as improper usage can exacerbate existing imbalances or introduce new biases. 
While synthetic data generation addresses class scarcity, its value hinges on how these samples interact with real data throughout the learning process.
To address this, we propose an RL-driven class adaptive sampler that dynamically calibrates synthetic-real data ratios during the classifier training. 
This dynamic adjustment ensures that synthetic data selectively compensates for the scarcities of real data while maintaining the authentic patterns that ensure feature fidelity.

Following the classical RL setting, we define a multi-agent $M=\{m_1,...,m_C\}$ with its current state $S$ that interacts with the environment $E$ by taking sequential actions $A$, aiming to maximize the expected reward.
Concretely, the environment $E$ is the breast US dataset containing both real and synthetic images. 
The state $S$ is the set of synthetic sample counts for $C$ classes in a mini-batch, $S=\{s_{1},s_{2},\ldots,s_{C}\}$. 
Each agent $m_i$ with its parameters $\theta_i$ adjusts the sampling number of synthetic images for a specific class.
The entire action space is defined as $A=\{a_{i}|i=1,2,\ldots,C\}$, where $a_i=[-2,0,+2]$ indicates the step size of the sampling number adjustment determined by probability $p(a_{i})$ output from $m_i(\theta_i)$.
As shown in Fig.~\ref{fig:online_samler}, the agents iterate $K$ episodes to generate $K$ different states by choosing different actions. 
$K$ identical classifiers are then trained in parallel using $K$ batch compositions to update their parameters, respectively. 
The classifier with the highest validation metric is selected, and its weights are used to initialize the classifier for the next epoch. 
After $K$ episodes, the parameters of the agents $\{\theta_i|i=1,2,\ldots,C\}$ are updated using the REINFORCE rule~\cite{williams1992simple}.
Particularly, the $i^{th}$ agent is updated following:
\begin{equation}
    \label{eq:1}
    \begin{aligned}
        \theta_{i}^{t+1} & =\theta_{i}^{t}+\eta\frac{1}{K}\sum_{j=1}^{K}(R_{j}-B^{t})\cdot\nabla_{\theta}\log(g(a^{t,j})) \\
        & =\theta_{i}^{t}+\eta\frac{1}{K}\sum_{j=1}^{K}(R_{j}-B^{t})\cdot\nabla_{\theta}\sum_{i=1}^{C}\log(p(a_{i}^{t,j})),
    \end{aligned}
\end{equation}
where $\eta$ is the learning rate, $g(a^{t,j})=\prod(p(a_{i}^{t,j}))$ represents the joint probability distribution of different actions in the $j^{th}$ episode at $t^{th}$ epoch.
$B^{t}$ is a baseline term to improve the stability of the agents~\cite{wang2022awsnet}, which is expressed as follows:
\begin{equation}
    \label{eq:2}
    B^{t}=(1-\gamma)\cdot B^{t-1}+\gamma\cdot(\frac{1}{K}\sum_{j=1}^{K}R_{j}),
\end{equation}
where $\gamma$ is a constant. Note that $B^{t}$ is simplified to $\gamma\cdot(\frac{1}{K}\sum_{j=1}^{K}R_{j})$ when $t=1$.
The reward $R_{j}$ is defined as $(\epsilon_{j}+0.04)^{3}$, where $\epsilon_{j}\in[0,1]$ is the specific metric calculated on the validation set with $j^{th}$ classifier. 
The cubic function is utilized to enhance the reward signal. 
This dynamic sampling adjustment automates mini-batch construction, effectively balancing long-tailed data distributions. 
Furthermore, it enables noise stabilization by modulating synthetic usage in sync with the model’s evolving discriminative capability, preventing excessive synthetic samples from corrupting learned embeddings.

\section{Experiments and Results}
\label{sec:Experimental Results}
\begin{table}[!t]
    \centering
    \caption{Performance comparison of different methods on the in-house Breast-LT-8 dataset and the public BreastMNIST dataset. Higher values mean better performance, except for FID. All metrics except FID and AUC are presented in percentages (\%).}
    \label{table:methods_comparison}
    \setlength{\tabcolsep}{0.49mm}
    \begin{tabular}{lccccccccccc}
    \toprule
    \multirow{2}[2]{*}{Methods} & \multicolumn{8}{c}{Breast-LT-8} & 
    \multicolumn{2}{c}{BreastMNIST~\cite{medmnistv2}} \\
    \cmidrule(lr){2-9} \cmidrule(lr){10-11}
    & F1 & Rec & Pre & All & Many & Med & Few & FID & \hspace{2.5mm}Acc & AUC \\
    \midrule
    Baseline & 30.94 & 30.81 & 31.11 & 70.13 & 80.68 & 21.29 & 0.00 & - & \hspace{2.5mm}84.20 & 0.866 \\
    CBFocal & 25.47 & 28.44 & 25.97 & 50.64 & 57.80 & 19.94 & 16.07 & - & \hspace{2.5mm}83.97 & 0.840 \\
    Logit-Adjust & 31.97 & 32.10 & 31.88 & 70.96 & 81.27 & 20.44 & 6.25 & - & \hspace{2.5mm}87.18 & 0.888 \\
    \midrule
    LIFT & 20.06 & 20.31 & 26.62 & 42.08 & 47.99 & 14.94 & 26.37 & - & \hspace{2.5mm}87.03 & 0.848 \\
    ProCo & 30.36 & 38.21 & 29.87 & 58.65 & 65.28 & 24.79 & 36.67 & - & \hspace{2.5mm}83.13 & 0.868 \\
    NCL & 29.15 & 35.10 & 29.44 & 54.78 & 61.73 & 19.69 & \textbf{39.29} & - & \hspace{2.5mm}88.46 & 0.861 \\
    \midrule
    GLMC & 34.06 & 38.83 & 32.57 & 66.22 & 75.29 & 26.61 & 26.79 & - & \hspace{2.5mm}84.62 & 0.857 \\
    MGDM & 32.78 & 33.41 & 32.58 & 70.00 & 80.11 & 20.52 & 12.50 & 36.28 & \hspace{2.5mm}87.05 & \textbf{0.951} \\
    Skin-SDM & 33.30 & 35.40 & 32.25 & 68.59 & 77.58 & 27.32 & 9.38 & 32.46 & \hspace{2.5mm}88.72 & 0.938 \\
    \textbf{Ours} & \textbf{35.23} & \textbf{38.98} & \textbf{34.39} & \textbf{72.31} & \textbf{82.14} & \textbf{28.26} & 31.25 & \textbf{30.19} & \hspace{2.5mm}\textbf{89.10} & 0.924 \\
    \bottomrule
    \end{tabular}
\end{table}
\begin{table}[ht]
    \centering
    \caption{Results of the ablation study using the Breast-LT-8 dataset.}
    \setlength{\tabcolsep}{1.5mm}
    \label{tabel:abation}
    \begin{tabular}{lccccccc}
    \toprule
    Methods & F1 & Rec & Pre & All & Many & Med & Few \\
    \midrule
    Baseline & 30.94 & 30.81 & 31.11 & 70.13 & 80.68 & 21.29 & 0.00       \\
    +SynClass& 31.82 & 32.14 & 32.33 & 70.58 & 80.42 & 19.39 & 9.38     \\
    +SynSketch & 34.73 & 34.89 &\textbf{34.94} & 72.18 & \textbf{82.23} & 21.70 & 12.95     \\
    +SynClass+Re-sampling& 29.48 & 34.96 & 29.35 & 60.58 & 67.81 & 20.17 & \textbf{31.70} \\ +SynClass+\R& 34.21 & 35.87 & 33.25 & 71.22 & 81.03 & 23.42& 15.62  \\     
    \textbf{Ours} & \textbf{35.23} & \textbf{38.98} & 34.39 & \textbf{72.31} & 82.14 & \textbf{28.26}& 31.25  \\
    \bottomrule
    \end{tabular}
\end{table}

\textbf{Datasets and implementation details.} 
We evaluated our method on two breast US datasets: an in-house long-tailed dataset Breast-LT-8 (max class imbalance ratio 47.98:1, see Fig.~\ref{fig:Intro Fig1}) and the public BreastMNIST dataset~\cite{medmnistv2} to provide complementary validation on class imbalance. 
Approved by the local IRB, the Breast-LT-8 dataset containing 5622 US images with 8 classes of different histological subtypes of breast lesions was collected from 2811 patients.
All images were resized to $256\times256$ and corresponding ground truth labels were obtained through biopsies.
The Breast-LT-8 was split randomly at the patient level with a ratio of 7:1:2 for training, validation, and testing. 
The BreastMNIST (containing 780 US images with size $224\times224$) was utilized for binary classification.
The synthesizer was trained using an AdamW optimizer with a learning rate of 1e-4 for 200 epochs.
The downsampling factor of VAE was 8.
We set the $\lambda$ in the sketch loss to 0.1.
For the~\R, the state was initialized to 2 and the episode $K=3$ in each epoch. 
The $\gamma$ was set to 0.99.
The agents were trained for 30 epochs using an Adam optimizer with a learning rate of 1e-3. 

\textbf{Method Comparison.} 
As shown in Table \ref{table:methods_comparison}, the proposed method was compared against both classical and the state-of-the-art (SOTA) approaches in long-tail classification, including: 
a) re-balancing-based, i.e., CBFocal \cite{cui2019classbalancedlossbasedeffective}, Logit-Adjust \cite{menon2020long}; 
b) architecture-enhanced, i.e., LIFT \cite{shi2024longtaillearningfoundationmodel}, ProCo \cite{du2024probabilistic}, NCL \cite{li2022nested}; 
c) augmentation-based, i.e., GLMC \cite{du2023global}, MGDM \cite{luo2024measurement}, Skin-SDM \cite{Parapat2024generating}.
Performance was calculated using common evaluation metrics such as F1-score (F1), precision (Pre), recall (Rec), all accuracy (All), shot accuracy (Many, Med, Few), Area Under Curve (AUC), as well as Fréchet Inception Distance (FID) for generative approaches. 
Note that we adopt ResNet with Balanced Softmax as a strong baseline for long-tailed classification to ensure fair benchmarking while the backbone can be easily replaced for future explorations. 

In \textbf{Breast-LT-8} dataset, compared to the re-weighting methods (rows 2-3), our approach demonstrated significant improvements across all metrics which indicates its capacity on effectively recognizing tail classes while preserving robust performance on head classes (e.g., F1=35.32, Many=82.14, Few=31.25). In terms of accuracy for many-shot and few-shot, CBFocal and Logit-Adjust exhibited opposite performance trends, highlighting that different re-weighting strategies distinctly prioritize different class samples. Interestingly, although architecture-enhanced methods (rows 4-6) showed promising improvements in recognizing tail classes compared to the baseline, these gains may not translate into significant benefits in F1-score, suggesting that excessive focus on the tail classes may hinder the overall performance (e.g., see the low F1-score=20.06 and All Acc=42.08 obtained by LIFT).  Overall, augmentation-based approaches (rows 7-10) achieved relatively higher F1-scores, likely arising from the increasing data quantity and diversity that mitigate data imbalance. Moreover, our proposed method outperformed other generative augmentation approaches (e.g., MGDM and Skin-SDM) across all metrics, validating its robust capability in adapting to long-tailed data distributions. Conversely, experimental results on the \textbf{BreastMNIST} dataset revealed narrowed performance gaps among mainstream methods. This convergence potentially originates from milder class imbalance and reduced task complexity in binary classification. Notably, our approach maintains superior accuracy, confirming its robustness for imbalanced classification scenarios. 

\textbf{Ablation studies.} 
To investigate the impact of each proposed component,  we conducted ablation experiments by ablating the generator(+Synclass), the structural perception branch(+SynSketch), and the RL-CAS sampler (see Table \ref{tabel:abation}). Note that we also implement a classical re-sampling approach following \cite{yang2023freemasksyntheticimagesdense} (see row 4) as an alternative for the RL-CAS sampler.
Compared to the baseline, the incorporation of synthetic data alone has augmented the dataset volume and diversity that lead to elevated F1-score and Rec (see row 1 and 2). Furthermore, the addition of the Sketch-Grounded Perception not only boosted the many- and medium-shot accuracy, but also further enhanced recognition of tail classes. These findings underscore the indispensable role of fine-grained structural guidance in synthesizing discriminative training samples. Meanwhile, the hand-crafted fixed resampling methods (row 4) yielded lower F1 score even with synthetic augmentation. This verifies the previous hypothesis that improper or excessive usage of synthetic samples may induce subtype-specific overfitting at the expense of holistic model capability. In contrast, the RL-CAS dynamically calibrates synthetic-real data ratios to fully exploit synthetic samples while preserving holistic classification accuracy. 

\section{Conclusion}
\label{sec:conclusion}
We propose a two-phase framework for long-tailed classification, addressing data asymmetry through high-fidelity synthesis and adaptive sampling. It integrates a reinforcement learning-driven sampler to balance head-class stability and tail-class exploration, alongside a sketch-grounded perception branch that injects anatomical priors to retain class-discriminative traits. Extensive experiments on in-house and public breast US datasets demonstrate balanced classification performance across different metrics. 
In the future, we will extend the framework to more long-tailed datasets and tasks. 

\begin{credits}
\subsubsection{Acknowledgments.}
This work was supported by the National Natural Science Foundation of China (No. 62471305, 62101342, 12326619, and 62171290); Guangdong Basic and Applied Basic Research Foundation (No.2025A1515011448, and 2023A1515012960); Science and Technology Planning Project of Guangdong Province (No. 2023A0505020002); Frontier Technology Development Program of Jiangsu Province (No. BF2024078).

\subsubsection{Disclosure of Interests.}
The authors have no competing interests to declare that are relevant to the content of this article.
\end{credits}

\bibliographystyle{splncs04}
\bibliography{Paper-5051}

\end{document}